\documentclass{article}

\usepackage[main, final]{template}
\usepackage[utf8]{inputenc} % allow utf-8 input
\usepackage[T1]{fontenc}    % use 8-bit T1 fonts
\usepackage{hyperref}       % hyperlinks
\usepackage{url}            % simple URL typesetting
\usepackage{booktabs}       % professional-quality tables
\usepackage{amsfonts}       % blackboard math symbols
\usepackage{nicefrac}       % compact symbols for 1/2, etc.
\usepackage{microtype}      % microtypography
\usepackage{xcolor}         % colors
\usepackage{graphicx}
\usepackage{amsmath}
\usepackage{wrapfig}

\title{Are More Tokens Rational? \\Inference-Time Scaling in Language Models as Adaptive Resource Rationality}

\author{%
  \textbf{Zhimin Hu}$^{1}$ \quad \textbf{Riya Roshan}$^{1}$ \quad \textbf{Sashank Varma}$^{1}$ \\
  \vspace{1ex} \\
  $^{1}$ Georgia Tech \\
}

\begin{document}

\maketitle

\begin{abstract}
Human reasoning is shaped by resource rationality -- optimizing performance under constraints. Recently, inference-time scaling has emerged as a powerful paradigm to improve the reasoning performance of Large Language Models by expanding test-time computation. Specifically, instruction-tuned (IT) models explicitly generate long reasoning steps during inference, whereas Large Reasoning Models (LRMs) are trained by reinforcement learning to discover reasoning paths that maximize accuracy. However, it remains unclear whether resource-rationality can emerge from such scaling without explicit reward related to computational costs. We introduce a Variable Attribution Task in which models infer which variables determine outcomes given candidate variables, input-output trials, and predefined logical functions. By varying the number of candidate variables and trials, we systematically manipulate task complexity. Both models exhibit a transition from brute-force to analytic strategies as complexity increases. IT models degrade on XOR and XNOR functions, whereas LRMs remain robust. These findings suggest that models can adjust their reasoning behavior in response to task complexity, even without explicit cost-based reward. This provides compelling evidence that resource rationality is an emergent property of inference-time scaling itself. \footnote{Code and data can be found at https://github.com/Paleuna/Inference-Time-Scale-Adaptation}
\end{abstract}

\section{Introduction}
What is intelligence, if not the art of doing well with less?  From biological brains to artificial agents, intelligent behavior requires--in fact, emerges from the efficient use of limited cognitive resources. This idea dates back to Herbert Simon’s theory of \emph{bounded rationality} \citep{simon1955behavioral}, which challenged the notion of 'perfect rationality' \citep{von1947theory} by highlighting the real-world -- and cognitive -- constraints of memory, computation, and time. Rather than seeking optimal solutions, agents `satisfice' by setting expectation thresholds and stopping once an acceptable option is found. This process of `constraint satisfaction' is now recognized as a fundamental component of reasoning and decision-making. Simon's insights have been developed across multiple disciplines. Within economics, Conlisk attempted to bring `deliberation costs' or `optimization costs' into the computation of utility \citep{CONLISK1988, CONLISK1996}. Within AI, \citet{BODDY1994} considered the design of agents planning solutions in time-constrained environments. In behavioral experiments, Payne and colleagues have studied the adaptivity of human decision making and problem solving under external constraints \citep{PAYNE1988, PAYNE1996}.

Expanding on Simon’s work, \citet{lieder2020resource} proposed that cognition is \emph{resource rational} and formalized reasoning as an optimization problem under cognitive constraints. In this view, agents seek to maximize the difference between expected utility and the cost of computation. \citet{callaway2022rational} extended this approach by modeling reasoning as a form of metacognitive control, where agents flexibly adapt between strategies based on their relative reward and processing cost. Such models of resource rationality frame strategy adaptation as a core computational property of bounded intelligence.

In this work, we aim to investigate whether resource rationality is not merely a descriptive account of human cognition, but a general computational principle of intelligent agents under constraint. While paradigms like Mouselab \citep{lieder2020resource, PAYNE1988} have been valuable for exploring how humans trade off information acquisition cost and accuracy--often shaped by preferences and heuristics--they focus primarily on \emph{external} sampling behavior. As such, they are less ideal for isolating the \emph{internal} computational trade-offs that underlie general strategy adaptation.

\begin{figure*}[t] 
  \centering
  \includegraphics[width=0.98\textwidth]{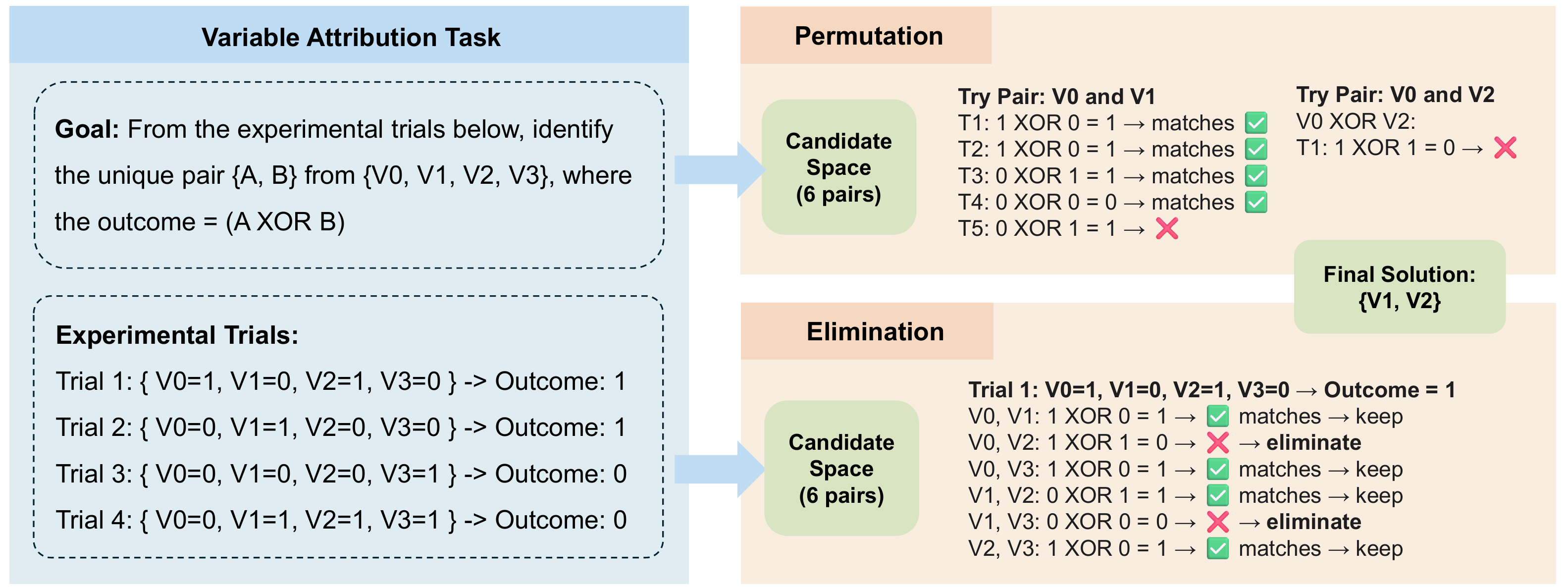} 
  \caption{Variable Attribution Task and its computational strategies. Given input–output records from $T$ trials and $N$ candidate variables, the model must identify the pair that determines the output via a latent Boolean function. Solving the task starts either from testing candidate pairs (permutation) or incrementally pruning inconsistent pairs (elimination).}
  \label{fig:overview}
\end{figure*}

To address this limitation, we introduce the Variable Attribution Task (VAT), a structured learning paradigm for investigating how agents manage internal resources. As shown in Figure~\ref{fig:overview}, the agent is given $N$ candidate variables and $T$ input–output trials where a predefined Boolean function maps a subset of the candidate variables to the outcome. This is a very general setup that can be mapped to multiple forms of reasoning, e.g., identifying causal variables in scientific reasoning. The VAT supports two cognitively distinct strategies: \emph{permutation}-based search, which performs serial hypothesis testing with low working memory demands; and \emph{elimination}-based inference, which maintains and prunes a hypothesis space based on disconfirming evidence and makes higher working memory demands. Both strategies operate over a hypothesis space that grows quadratically with the number of candidate variables and evaluate it across $T$ trials in the worst case, which results in similar worst-case complexity $O(T \times N^2)$. However, their convergence efficiency diverges depending on the structure of the target function. Under AND-like functions, elimination can rapidly reduce the candidate space, yielding faster identification of the subset of relevant variables. In contrast, under XOR-like functions, disconfirmation typically offers less pruning power.

A recent shift in the reasoning paradigm of Large Language Models (LLMs) has made them useful testbeds for the question of whether resource rationality can emerge naturally, as a computational principle of optimization under constraint. This shift leads to inference time scaling, which attempts to improve reasoning by expanding the computation at test time. Notably, Chain-of-Thought (CoT) \citep{wei2022chain} demonstrated that generating reasoning steps boosts performance without additional training. Instruction-tuned (IT) models build on this by undergoing Supervised Fine-Tuning (SFT) on CoT-style data, while Large Reasoning Models (LRMs) such as DeepSeek R1 \citep{guo2025deepseek} further optimize reasoning behavior through Reinforcement Learning (RL). While prior work evaluating human and machine reasoning by their ability to produce fixed answers \citep{webb2023emergent, hu2024failures} and short-form responses \citep{kabra2025modeling}, inference-time scaling enables models to generate explicit reasoning trajectories. These reasoning tokens serve as a traceable cognitive process, allowing us to analyze how models perform strategy adaptation over the course of inference.

Use the VAT, we systematically manipulate task complexity to test IT models and LRM, and track their shifting between the permutation and elimination strategies. Unlike paradigms that emphasize exploration cost, VAT directly targets strategy adaptation, offering a cleaner view on how computational effort is reallocated under constraint. Our results show that both IT models and LRMs exhibit a graded transition from permutation to elimination as task complexity increases. However, for the XOR and XNOR functions--where pruning offers little benefit and maintaining the hypothesis space becomes costly--this transition is diminished or absent. Furthermore, while both model types show similar patterns of strategy adaptation, their performance diverges: LRMs maintain stable accuracy across functions, whereas IT models degrade for the more-demanding XOR and XNOR functions. Our findings demonstrate that resource rationality is an emergent property of inference-time scaling, rather than a behavior requiring explicit cost-based rewards. We observed that both IT models and LRMs autonomously adapt their computational strategies to task structures, suggesting that resource-rational behavior arises inherently from optimization under finite capacity. Consequently, resource rationality may serve as a computational principle of how complex systems reorganize computation to navigate environmental and internal constraints.

\section{Related Work}

\subsection{Inference Time Scaling}

Inference-time scaling refers to the paradigm where models improve performance by allocating more computation at test time. Early work relied on prompt-based techniques. CoT \citep{wei2022chain} instructs models to generate intermediate reasoning steps before producing the final answer, while ToT \citep{yao2023tree} frames inference as an explicit search process with lookahead and backtracking.

Recent approaches aim to integrate such intermediate computation into post-training. Models are first fine-tuned on long CoT datasets via SFT, enabling instruction-tuned (IT) models to follow prompts while generating reasoning steps. In contrast, reasoning models explicitly separate thinking tokens from output tokens and are further trained with RL to regulate reasoning. In DeepSeek R1 \citep{guo2025deepseek}, RL rewards correct final answers, encouraging reasoning as latent search. In Qwen3 \citep{yang2025qwen3}, RL focuses on reasoning coherence and control, penalizing both inconsistencies and incorrect answers. Empirically, increasing inference-time yields consistent performance gains \citep{snell2025scaling}.

\subsection{Induction and Attribution}

Induction is the learning of latent regularities from finite observations. The processing of inferring specific causes of observed outcomes, termed as attribution, forms the essential part of induction \citep{cheng1990probabilistic}. In human induction, \citet{cheng1990probabilistic} suggested that causal strength is judged by comparing the probability of an outcome in the presence versus the absence of a potential cause, and this contrastive process is, in essence, a form of attribution. However, the exponential (in the number of candidates) size of the hypothesis space forces biological agents to rely on local updates and sequential hypothesis testing rather than global optimization \citep{bramley2017constructing}. Attribution difficulty also depends on the logic of the causal relation. Humans favor natural configural strategies (e.g., conjunctive logic) over non-monotonic forms \citep{ganzach1995learning}. To manage cost-accuracy tradeoffs, distinct strategies emerge. \citet{varma2018middle} found that students use either instant rationality for rapid pruning or delayed rationality that tolerates inconsistency for global evaluation, independent of general cognitive ability.

\section{Methods}

\subsection{Task Design and Logic Space} 

The VAT is designed to test a model's ability to perform causal induction under varying computational constraints. Given $N$ candidate variables $\{V_0, V_1, \dots, V_{N-1}\}$ and $T$ experimental trials, the model must identify a unique pair of active variables $(V_i, V_j)$ that determines the system output $Y$ according to a predefined logical function $f$. We consider the complete space of bivariate Boolean functions. There are $2^{2^2} = 16$ possible functions for two inputs. However, we exclude the 2 constant functions and the 4 univariate functions (which depend on only one or zero variables) because they do not require a variable pair and are thus much simpler to determine. The remaining 10 non-trivial functions are categorized into three groups based on their truth-table characteristics:
\begin{itemize}
\item \textbf{Conjunctive:} Functions where only 1 of the 4 input combinations yields a positive outcome (e.g., $A \land (\lnot B)$).
\item \textbf{Disjunctive:} Functions where 3 of the 4 combinations yield a positive outcome, or conversely, only 1 yields a negative outcome (e.g., $(\lnot A) \lor B$).
\item \textbf{XOR/XNOR:} Non-linear functions where the outcomes are balanced (2 positive, 2 negative). 
\end{itemize}

\subsection{Information Ratio}

To systematically manipulate task complexity, we define the \textbf{Information Ratio} ($\rho$) as the relationship between the available experimental information and the hypothesis space: $\rho = {2^T} /{\binom{N}{2}}$, where $T$ is the number of trials and $\binom{N}{2}$ represents the total possible combinations of active variables. When $\rho$ is small (near the theoretical minimum $1$), the experimental trials are extremely dense, such that almost every trial is necessary to eliminate incorrect hypotheses, and vice versa.

\subsection{Computational Strategies}

The VAT is, in principle, an NP problem: although a solution can be verified in polynomial time, no known algorithm can solve it in polynomial time in the general case. Theoretically, the worst-case complexity is $O(T \times N^2)$. Solving the task requires evaluating candidate pairs of variables against the observed input–output mappings. It is naturally done via two strategies. The permutation strategy iterates over all $\binom{N}{2}$ candidate pairs and checks their consistency across all trials. The elimination strategy processes each trial to incrementally prune incompatible pairs. To preclude heuristic shortcuts, we ensured that the input distributions (i.e., the frequency of $0$s and $1$s) for candidate variables were similar to those of the ground-truth variables. This prevents models from exploiting statistical cues (e.g., “$V_1$ is only one with a complete set of $1$ and $0$) and forces reliance on logical reasoning. While both strategies have similar theoretical complexity, they differ in cognitive demands and empirical efficiency; see Figure~\ref{fig:overview}. The permutation strategy places a relatively low burden on working memory, as only one hypothesis needs to be maintained and tested at a time. In contrast, the elimination strategy requires updating and maintaining a global hypothesis space, imposing a higher cognitive load.

\subsection{Materials}

To systematically investigate model behavior under varying task complexity, we generated a total of 3000 variable attribution samples spanning 10 logical functions. We considered 10 values of $N$, the number of candidate variables: $[N \in \{3, 4, 5, 6, 7, 8, 10, 12, 14, 16\}$. For each $N$, we computed the minimal number of trials $T_{\min}$ required to uniquely identify the correct variable pair. Thus, we adjusted $T_{\min}$ to: $T = T_{\min} + \{0,1,2,3,4,5\}$. For each combination of $(N, T)$, we randomly generated 5 samples, which results in $300$ samples per logical function. 

\begin{figure*}[t] 
  \centering
  \includegraphics[width=1\textwidth]{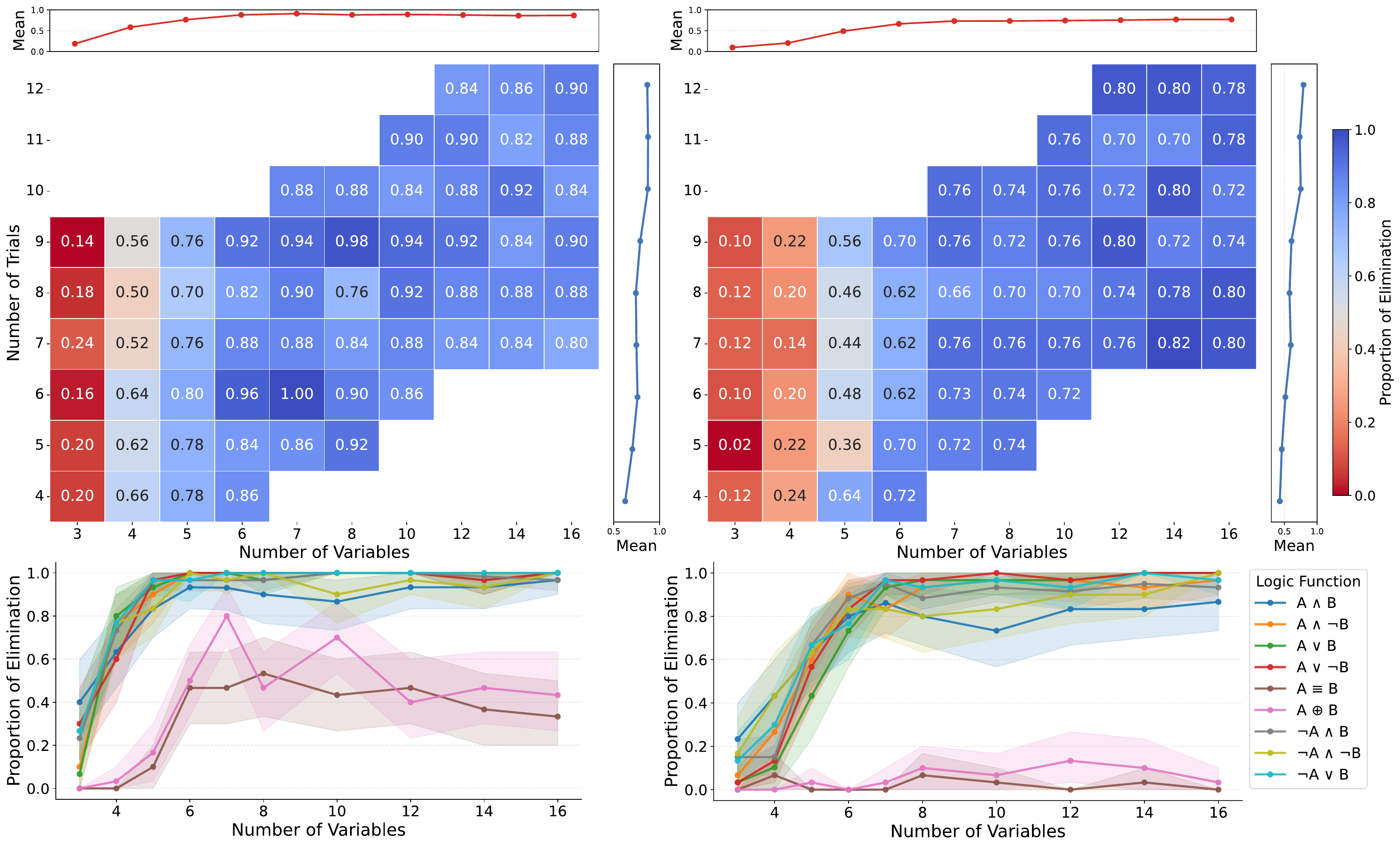} 
  \caption{Strategy distribution across task complexity. (Top) Heatmaps showing the proportion of elimination strategies along the $N$ and $T$, respectively, where the top and right marginal plots represent the mean proportion trends. (Bottom) Proportion of elimination for specific logical functions. (Left) Deepseek-R1. (Right) Qwen3-thinking.}
  \label{fig:main_strategy_result}
\end{figure*}

\subsection{Models}

We evaluated two families of LLMs on this dataset: 1) \texttt{DeepSeek R1 \& V3}; 2) \texttt{Qwen3-Next-80B-A3B-Thinking \& Instruct} (hereafter referred to as Qwen3-Thinking \& Instruct). All 4 models were evaluated using the same base prompt. For the thinking models, we analyzed their ``thinking'' tokens to identify the underlying computational strategies. Token usage is measured using character count. For the thinking models, we included both the reasoning trace and the final output in this count. For Qwen3-Instruct, we additionally tested a constrained setting in which the model was instructed to output only the final answer, which serves as the baseline for accuracy.

\subsection{LLM as Judge}

To identify the reasoning strategies employed by models, we used a high-capacity language model, \texttt{Kimi-K2-Instruct-0905}, as an external judge. Given a model’s response, the judge was instructed to classify the strategy as permutation, elimination, or invalid. To ensure the validity of this approach, we constructed a human-labeled evaluation set. We evenly sampled responses from Qwen3-Thinking and DeepSeek-R1, manually annotated their strategies, and curated a balanced subset of 100 examples with roughly equal numbers of permutation and elimination cases. We then evaluated the prompt on this set. The LLM-based judge achieved an accuracy of 0.86 and a Cohen’s Kappa of 0.76, indicating substantial agreement with human annotations and validating its use for analysis.

\section{Results}

\begin{wrapfigure}{l}{0.5\textwidth} 
  \centering
  \vspace{-10pt} 
  \includegraphics[width=0.5\textwidth]{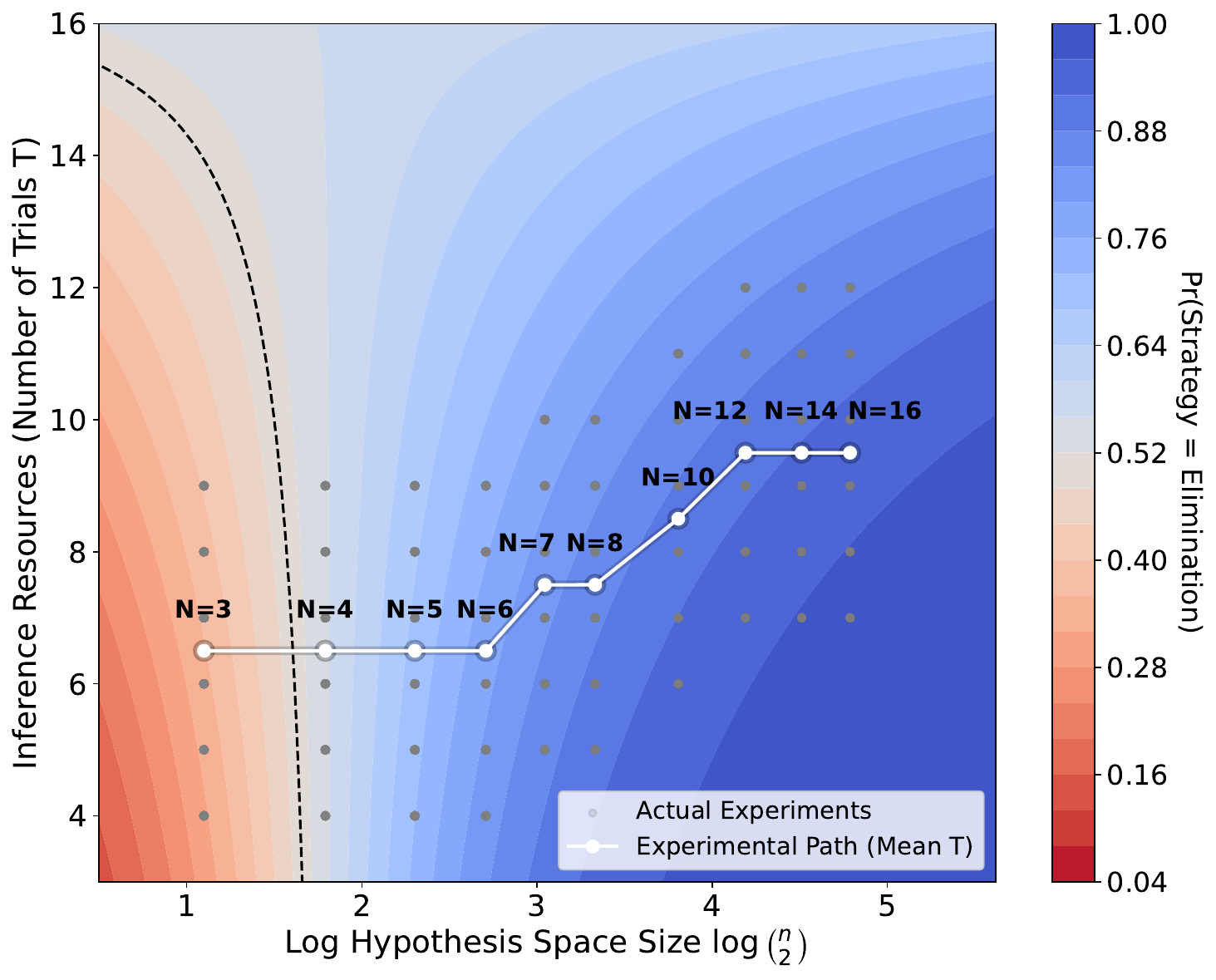}
  \caption{The fitted decision landscape of strategy selection. The heatmap represents the predicted probability of choosing the elimination strategy based on the interaction model. The gray dots represent actual experimental samples, and the white line denotes the mean experimental path ($T_{mean}$ for each $N$). The dashed line indicates the 50\% decision boundary.}
  \label{fig:landscape}
  \vspace{-15pt} 
\end{wrapfigure}

\subsection{Emergent Strategy Selection}
We first investigated whether LLMs adaptively switch between reasoning strategies. As shown in the upper row of Figure~\ref{fig:main_strategy_result}, we observed a clear phase transition in strategy selection for both the DeepSeek and Qwen families. When the number of candidate variables $N$ is small (e.g., $N < 4$), models predominantly employ a permutation strategy, iterating through possible pairs to check consistency. However, as $N$ increases, the hypothesis space $\binom{N}{2}$ grows quadratically. The models adapt and transition toward an elimination strategy, which processes trials one-at-a-time to prune the search space. This trend is observed across both Large Reasoning Models (LRMs) and their instruction-tuned (IT) counterparts (DeepSeek V3 and Qwen3-Instruct). However, the relationship between the proportion of elimination and the $T$ is unclear since the data is unbalanced. In the next section, we will fit an interaction model to systematically investigate how task complexity shapes strategy selection.

Notably, we observed a divergence in strategy selection for the XOR and XNOR functions in the lower row of Figure~\ref{fig:main_strategy_result}. While DeepSeek models show a moderate shift toward elimination for these functions, the proportion remains lower than for conjunctive/disjunctive tasks. In contrast, the Qwen family almost entirely eschews elimination for XOR-like logical functions, favoring permutation even as $N$ increases. From a Resource Rationality perspective, this divergence is highly instructive. In conjunctive logic (e.g., AND), a single positive trial can eliminate a vast region of the hypothesis space, making the ``working memory'' cost of elimination highly efficient. By contrast, the XOR/XNOR functions are non-monotonic and "information-dense"; they do not allow for rapid pruning of the variable space through single observations. If the computational cost of maintaining a very large and only slowly shrinking hypothesis space exceeds the cost of a linear permutation search, remaining in the permutation regime is the resource-optimal choice. The fact that the models modulate their strategy based on the prunability of the logic--rather than just the raw size of $N$--suggests they are sensitive to the difference between expected utility and cost.

\begin{table}[t]
\centering
\fontsize{8.5pt}{9pt}
\begin{tabular}{lccc}
\toprule
\textbf{Model} & \textbf{AIC} & \textbf{Pseudo $R^2$} & \textbf{Wald P-values}\\
\midrule
\rule{0pt}{2.5ex}Elimination $\sim \log \binom{N}{2} \times T$
& 2727.29 & 0.160 & $<$ 0.001 (all) \\

\rule{0pt}{2.5ex}Elimination $\sim \log \binom{N}{2} + T$
& 2751.10 & 0.152 & $<$ 0.001 (all)\\

\rule{0pt}{2.5ex}Elimination $\sim \log \binom{N}{2}$
& 2765.76 & 0.147 & $<$ 0.001\\

\rule{0pt}{2.5ex}Elimination $\sim$ $\rho$
& 3138.23 & 0.032 & $<$ 0.001\\

\rule{0pt}{2.5ex}Elimination $\sim T$
& 3178.55 & 0.019 & $<$ 0.001\\
\bottomrule
\end{tabular}
\caption{Model comparison for predicting the frequency of applying the elimination strategy. $\log \binom{N}{2}$ denotes hypothesis space size, $T$ denotes the number of trials, and $\rho$ denotes the information ratio. \textit{all} indicates that the $\log \binom{N}{2}$ and $T$, and their interaction all have a p-value below 0.001.}
\label{tab:model_comparison}
\end{table}

% \begin{figure}[h]
%   \centering
%   \includegraphics[width=0.48\textwidth]{figure/landscape.png} 
%   \caption{The fitted decision landscape of strategy selection. The heatmap represents the predicted probability of choosing the elimination strategy based on the interaction model. The gray dots represent actual experimental samples, and the white line denotes the mean experimental path ($T_{mean}$ for each $N$). The dashed line indicates the 50\% decision boundary.}
%   \label{fig:landscape}
% \end{figure}

\subsection{Sensitivity to Task Complexity}
We next examine which components of task complexity drive this shift in strategy selection. To prevent multicollinearity, we centered the predictors before performing a logistic regression to quantify the drivers of strategy selection. As shown in Table~\ref{tab:model_comparison}, the interaction model--incorporating both the log-hypothesis space size and the number of trials--yielded the best fit ($AIC = 2727.29, \text{Pseudo } R^2 = 0.160$).

We used the interaction model's coefficients to visualize the decision landscape. As shown in Figure~\ref{fig:landscape}, two competing forces drive the probability (in terms of frequency) of choosing elimination. First, as the size of the hypothesis space increases, the model is more likely to adopt elimination. This is supported by a strong positive coefficient ($\beta = 1.0212, p < .001$) on the logarithm of hypothesis space sizes. At the same time, the number of trials has the opposite effect. When more trials are available, the model tends to slightly fall back on the simpler permutation strategy--even when the hypothesis space is large. This is reflected in the negative coefficient for trial count ($\beta = -0.1466, p < .001$).

In Figure~\ref{fig:landscape}, we observed that elimination is preferred only when pruning substantially reduces the hypothesis space--typically when $N$ is large, and $T$ is limited. As $T$ increases, the marginal value of each additional observation drops: the model can rely on permutation without incurring high error, making the added cost of elimination unjustified. In resource-rational terms, when $\mathbb{E}[\text{gain from pruning}] < \text{cost}_{\text{elimination}}$, the optimal choice shifts back to permutation. The landscape thus captures a dynamic cost-utility boundary, where strategy selection is sensitive not just to task size, but to how informative each trial is for pruning.

\begin{figure*}[h]
  \centering
  \includegraphics[width=1\textwidth]{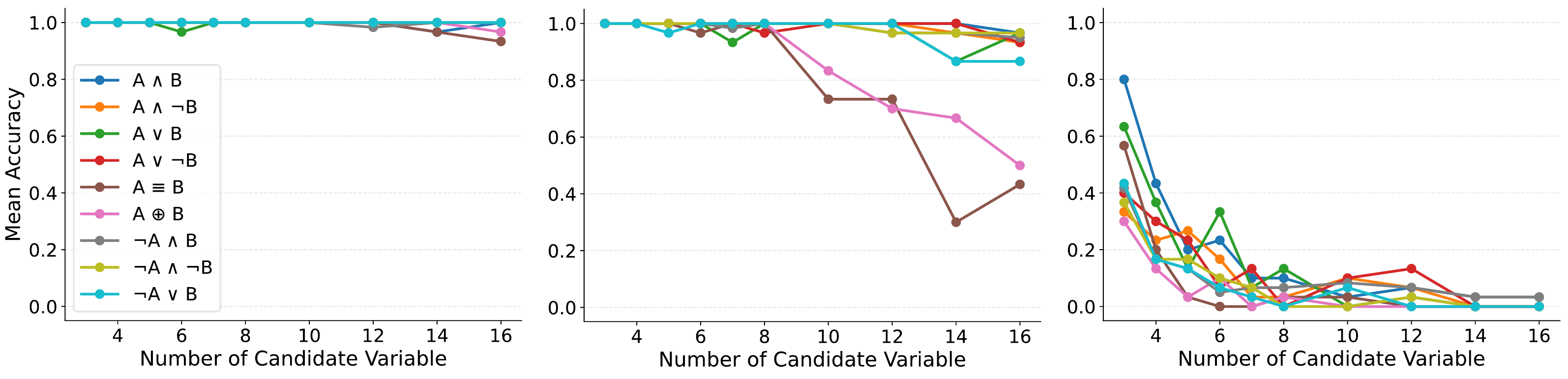} 
  \caption{Mean accuracy across logical functions as task complexity ($N$) increases (DeepSeek). From top to bottom: DeepSeek R1, V3, and V3 with direct answer.}
  \label{fig:accuracy}
\end{figure*}

\subsection{Accuracy and Reasoning Effort}
To evaluate how accuracy scales with reasoning effort, we compare the performance and character consumption (as a proxy for effort) of IT models and their LRM counterparts against a baseline of IT models constrained to direct-answer responses.

As depicted in Figure~\ref{fig:accuracy}, LRMs exhibit remarkable robustness, maintaining near-perfect accuracy regardless of logical function. In contrast, IT models show a performance drop on XOR/XNOR tasks as $N$ increases. The direct answer baseline fails almost immediately, with accuracy quickly approaching chance levels or zero. This confirms that Inference-time Scaling is the mechanism that allows models to boost reasoning. Meanwhile, the extended reasoning chain in LRMs is not merely a verbose version of standard output; rather, it allows the model to stabilize latent representations of non-linear relationships (like XOR) that are otherwise lost in the shallow processing of IT models.

\begin{figure*}[h]
  \centering
  \includegraphics[width=1\textwidth]{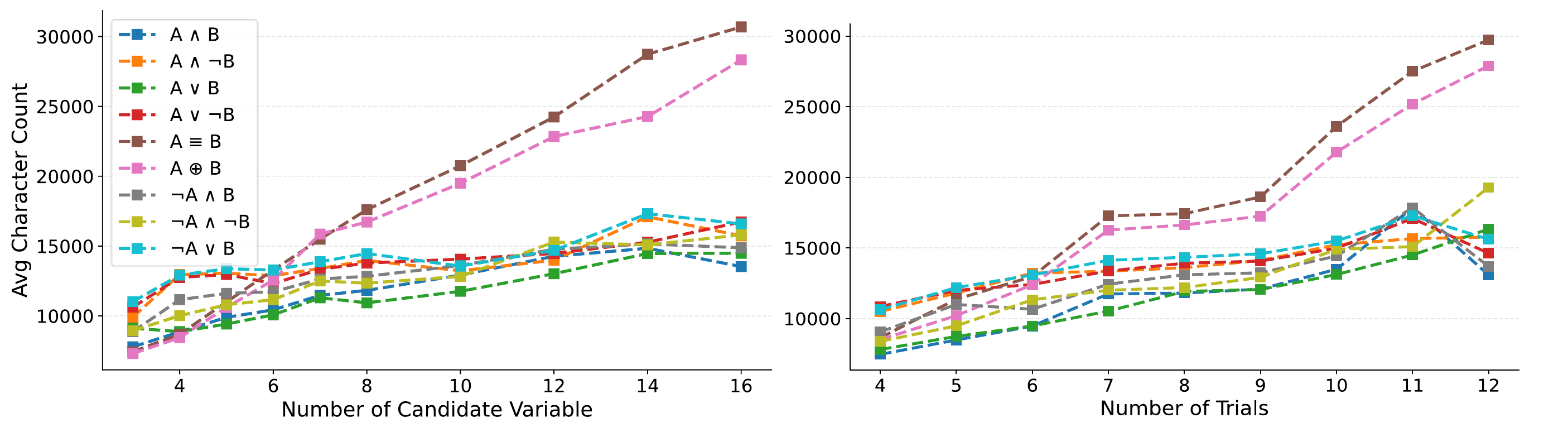} 
  \caption{Computational scaling behavior (DeepSeek R1). (Left) Average character count relative to $N$. (Right) Scaling relative to trials $T$. The dashed lines (XOR/XNOR) illustrate a higher intercept and steeper slope, indicating that logically complex tasks demand more inference resources.}
  \label{fig:scaling_char}
\end{figure*}

Meanwhile, as shown in Figure~\ref{fig:scaling_char}, character counts scale linearly with both $N$ and $T$ across all logic types. Notably, XOR and XNOR functions consistently require higher character counts, reflecting the inherent computational complexity of non-monotonic attribution. These phenomena suggest that models dynamically allocate more computational time as a trade-off to stabilize latent representations of non-linear relationships. This result potentially serves as a behavioral analog for human response time (RT), offering a predictive benchmark for future human study.

\section{Discussion}

A common critique of LLMs is that their apparent reasoning abilities reflect probabilistic pattern matching rather than principled inference \citep{kambhampati2024can}. From this perspective, model behavior is driven primarily by surface-level retrieval. Our findings challenge this view by showing that inference-time scaling functions as an implicit mechanism for resource-rational behavior that adaptively allocates computational resources in response to various task demands. In particular, IT models and LRMs exhibit adaptive selection of computational strategies as task demands increase. Notably, this adaptation emerges without any explicit training objective that penalizes or rewards computational cost, and appears largely independent of the specific post-training procedure. These results suggest that resource rationality may arise as an emergent property of large-scale optimization under constraint, rather than requiring explicit normative enforcement.

This analysis supports a broader interpretation of resource rationality. Rather than viewing it solely as a prescriptive framework that specifies how agents \emph{should} allocate resources, resource rationality may also serve as a descriptive lens for understanding how complex systems tend to behave when operating under finite capacity. In this sense, inference-time scaling does not merely improve accuracy, but reorganizes computation in a way that reflects sensitivity to both environmental structures and internal limitations.

These observations also resonate with longstanding findings in cognitive neuroscience. Prior work has emphasized that human cognition is constrained by available cortical resources, and that increasing task demands are accompanied by the recruitment of additional neural circuitry \citep{just2007organization, varma2014subjective}. In our setting, reasoning tokens can be interpreted as an externalized workspace that supports the maintenance of intermediate states during inference. When models confront high-dimensional hypothesis spaces, they generate longer reasoning traces to preserve partial eliminations and candidate relations. While this behavior is reminiscent of how the human brain increases prefrontal engagement under high working-memory load \citep{NORMAN1975, NORMAN1986}, we emphasize that this analogy remains tentative. An alternative explanation is that reasoning length reflects limitations in attention-based state maintenance rather than deliberate resource allocation. Distinguishing between these possibilities requires further investigation into the interaction between attention capacity, representation stability, and inference-time computation.

\bibliographystyle{unsrtnat} % 或 neurips 专用格式
\bibliography{bibliography}

@article{wei2022chain,
  title={Chain-of-thought prompting elicits reasoning in large language models},
  author={Wei, Jason and Wang, Xuezhi and Schuurmans, Dale and Bosma, Maarten and Xia, Fei and Chi, Ed and Le, Quoc V and Zhou, Denny and others},
  journal={Advances in neural information processing systems},
  volume={35},
  pages={24824--24837},
  year={2022}
}

@article{yao2023tree,
  title={Tree of thoughts: Deliberate problem solving with large language models},
  author={Yao, Shunyu and Yu, Dian and Zhao, Jeffrey and Shafran, Izhak and Griffiths, Tom and Cao, Yuan and Narasimhan, Karthik},
  journal={Advances in neural information processing systems},
  volume={36},
  pages={11809--11822},
  year={2023}
}

@article{lieder2020resource,
  title={Resource-rational analysis: Understanding human cognition as the optimal use of limited computational resources},
  author={Lieder, Falk and Griffiths, Thomas L},
  journal={Behavioral and brain sciences},
  volume={43},
  pages={e1},
  year={2020},
  publisher={Cambridge University Press}
}

@article{guo2025deepseek,
  title={Deepseek-r1 incentivizes reasoning in llms through reinforcement learning},
  author={Guo, Daya and Yang, Dejian and Zhang, Haowei and Song, Junxiao and Wang, Peiyi and Zhu, Qihao and Xu, Runxin and Zhang, Ruoyu and Ma, Shirong and Bi, Xiao and others},
  journal={Nature},
  volume={645},
  number={8081},
  pages={633--638},
  year={2025},
  publisher={Nature Publishing Group UK London}
}

@article{simon1955behavioral,
  title={A behavioral model of rational choice},
  author={Simon, Herbert A},
  journal={The quarterly journal of economics},
  pages={99--118},
  year={1955},
  publisher={JSTOR}
}

@article{callaway2022rational,
  title={Rational use of cognitive resources in human planning},
  author={Callaway, Frederick and van Opheusden, Bas and Gul, Sayan and Das, Priyam and Krueger, Paul M and Griffiths, Thomas L and Lieder, Falk},
  journal={Nature human behaviour},
  volume={6},
  number={8},
  pages={1112--1125},
  year={2022},
  publisher={Nature Publishing Group UK London}
}

@inproceedings{hu2024failures,
  title={Failures and successes to learn a core conceptual distinction from the statistics of language},
  author={Hu, Zhimin and van Paridon, JP and Lupyan, Gary},
  year={2024},
  organization={The Evolution of Language: Proceedings of the 15th International Conference~…}
}

@inproceedings{kabra2025modeling,
  title={Modeling Understanding of Story-Based Analogies Using Large Language Models},
  author={Kabra, Keshav and Inani, Kalit and Marupudi, Vijay and Varma, Sashank},
  booktitle={Proceedings of the Annual Meeting of the Cognitive Science Society},
  volume={47},
  year={2025}
}

@article{webb2023emergent,
  title={Emergent analogical reasoning in large language models},
  author={Webb, Taylor and Holyoak, Keith J and Lu, Hongjing},
  journal={Nature Human Behaviour},
  volume={7},
  number={9},
  pages={1526--1541},
  year={2023},
  publisher={Nature Publishing Group UK London}
}

@article{kambhampati2024can,
  title={Can large language models reason and plan?},
  author={Kambhampati, Subbarao},
  journal={Annals of the New York Academy of Sciences},
  volume={1534},
  number={1},
  pages={15--18},
  year={2024},
  publisher={Wiley Online Library}
}

@inproceedings{snell2025scaling,
  title={Scaling LLM test-time compute optimally can be more effective than scaling parameters for reasoning},
  author={Snell, Charlie Victor and Lee, Jaehoon and Xu, Kelvin and Kumar, Aviral},
  booktitle={International Conference on Learning Representations},
  year={2025}
}

@article{yang2025qwen3,
  title={Qwen3 technical report},
  author={Yang, An and Li, Anfeng and Yang, Baosong and Zhang, Beichen and Hui, Binyuan and Zheng, Bo and Yu, Bowen and Gao, Chang and Huang, Chengen and Lv, Chenxu and others},
  journal={arXiv preprint arXiv:2505.09388},
  year={2025}
}

@article{just2007organization,
  title={The organization of thinking: What functional brain imaging reveals about the neuroarchitecture of complex cognition},
  author={Just, Marcel Adam and Varma, Sashank},
  journal={Cognitive, Affective, \& Behavioral Neuroscience},
  volume={7},
  number={3},
  pages={153--191},
  year={2007},
  publisher={Springer}
}

@article{varma2014subjective,
  title={The subjective meaning of cognitive architecture: A Marrian analysis},
  author={Varma, Sashank},
  journal={Frontiers in psychology},
  volume={5},
  pages={440},
  year={2014},
  publisher={Frontiers Media SA}
}

@article{von1947theory,
  title={Theory of games and economic behavior, 2nd rev},
  author={Von Neumann, John and Morgenstern, Oskar},
  year={1947},
  publisher={Princeton university press}
}

@phdthesis{bramley2017constructing,
  title={Constructing the world: Active causal learning in cognition},
  author={Bramley, Neil Robert},
  year={2017},
  school={UCL (University College London)}
}

@article{ganzach1995learning,
  title={The learning of natural configural strategies},
  author={Ganzach, Yoav and Czaczkes, Benjamin},
  journal={Organizational Behavior and Human Decision Processes},
  volume={63},
  number={2},
  pages={195--206},
  year={1995},
  publisher={Elsevier}
}

@article{varma2018middle,
  title={Middle school students' approaches to reasoning about disconfirming evidence},
  author={Varma, Keisha and Boekel, Martin Van and Varma, Sashank},
  journal={Journal of educational and developmental psychology},
  volume={8},
  number={1},
  pages={28--42},
  year={2018},
  publisher={Canadian Center of Science and Education.}
}

@article{CONLISK1988,
  title={Optimization cost},
  author={Conlisk, John},
  journal={Journal of Economic Behavior \& Organization},
  volume={9},
  number={3},
  pages={213--228},
  year={1988},
  publisher={Elsevier}
}

@article{CONLISK1996,
  title={Why bounded rationality?},
  author={Conlisk, John},
  journal={Journal of economic literature},
  volume={34},
  number={2},
  pages={669--700},
  year={1996},
  publisher={JSTOR}
}

@article{BODDY1994,
  title={Deliberation scheduling for problem solving in time-constrained environments},
  author={Boddy, Mark and Dean, Thomas L},
  journal={Artificial Intelligence},
  volume={67},
  number={2},
  pages={245--285},
  year={1994},
  publisher={Elsevier}
}

@article{PAYNE1988,
  title={Adaptive strategy selection in decision making.},
  author={Payne, John W and Bettman, James R and Johnson, Eric J},
  journal={Journal of experimental psychology: Learning, Memory, and Cognition},
  volume={14},
  number={3},
  pages={534},
  year={1988},
  publisher={American Psychological Association}
}

@article{PAYNE1996,
  title={When time is money: Decision behavior under opportunity-cost time pressure},
  author={Payne, John W and Bettman, James R and Luce, Mary Frances},
  journal={Organizational behavior and human decision processes},
  volume={66},
  number={2},
  pages={131--152},
  year={1996},
  publisher={Elsevier}
}

@article{cheng1990probabilistic,
  title={A probabilistic contrast model of causal induction.},
  author={Cheng, Patricia W and Novick, Laura R},
  journal={Journal of personality and social psychology},
  volume={58},
  number={4},
  pages={545},
  year={1990},
  publisher={American Psychological Association}
}

@incollection{NORMAN1986,
  title={Attention to action: Willed and automatic control of behavior},
  author={Norman, Donald A and Shallice, Tim},
  booktitle={Consciousness and self-regulation: Advances in research and theory volume 4},
  pages={1--18},
  year={1986},
  publisher={Springer}
}

@article{NORMAN1975,
  title={On data-limited and resource-limited processes},
  author={Norman, Donald A and Bobrow, Daniel G},
  journal={Cognitive psychology},
  volume={7},
  number={1},
  pages={44--64},
  year={1975},
  publisher={Elsevier}
}

\end{document}